\documentclass[10pt,twocolumn,letterpaper]{article}

\usepackage{cvpr}              

\usepackage[dvipsnames]{xcolor}

\definecolor{cvprblue}{rgb}{0.21,0.49,0.74}
\usepackage[pagebackref,breaklinks,colorlinks,citecolor=cvprblue]{hyperref}
\usepackage{extra_styles}
\usepackage{graphicx}
\usepackage{multirow}
\usepackage{textpos}
\usepackage[accsupp]{axessibility}

\newcommand\nonumberfootnote[1]{%
  \begingroup%
  \let\thefootnote\relax%
  \footnotetext{#1}%
  \addtocounter{footnote}{-1}%
  \endgroup%
}
\def\paperID{11816} 
\def\confName{CVPR}
\def\confYear{2024}

\title{Faces that Speak: \\Jointly Synthesising Talking Face and Speech from Text}

\author{Youngjoon Jang\(^{1*}\) \quad Ji-Hoon Kim\(^{1*}\) \quad Junseok Ahn\(^1\) \quad Doyeop Kwak\(^1\) \\ Hong-Sun Yang\(^2\) \quad Yoon-Cheol Ju\(^2\) \quad Il-Hwan Kim\(^2\) \quad Byeong-Yeol Kim\(^2\) \quad Joon Son Chung\(^1\) \\ 
$^{1}$Korea Advanced Institute of Science and Technology, $^2$42dot Inc., Republic of Korea}

\begin{document}

\twocolumn[{
\maketitle
\begin{center}
    \captionsetup{type=figure}
    \includegraphics[width=0.85\textwidth]{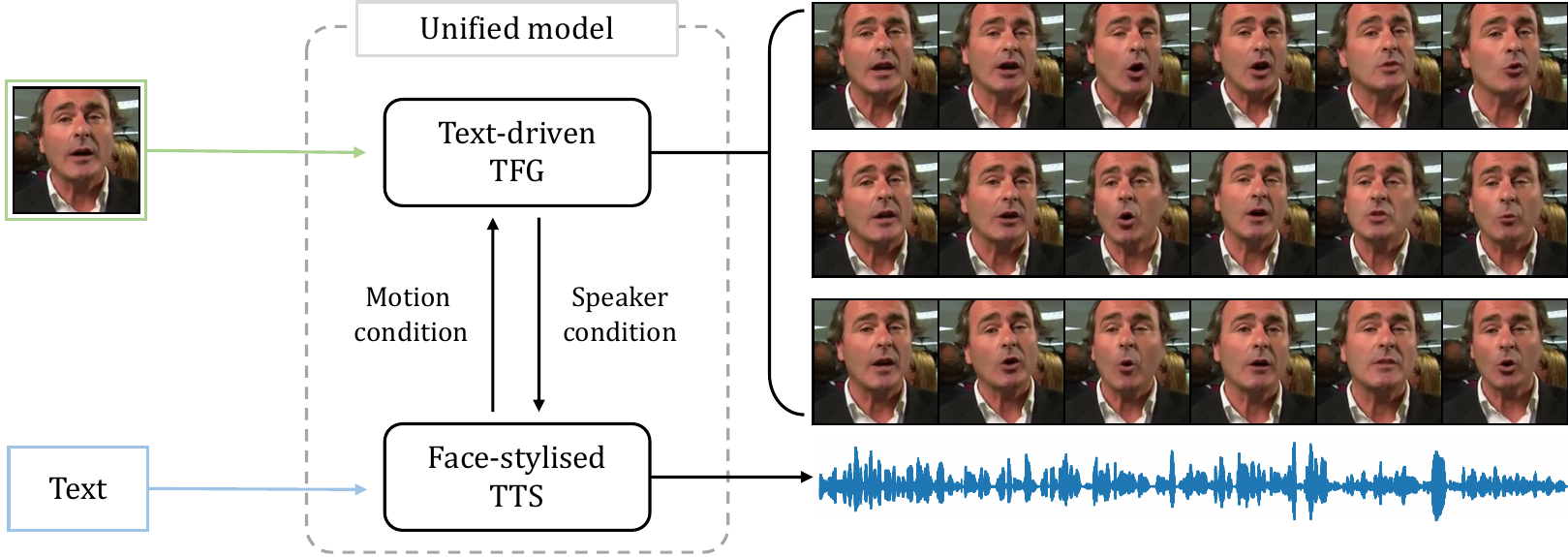}
    \vspace{-2.0mm}
    \captionof{figure}{Our framework integrates Talking Face Generation (TFG) and Text-to-Speech (TTS) systems, generating synchronised natural speech and a talking face video from a single portrait and text input. 
    Our model is capable of variational motion generation by conditioning the TFG model with the intermediate representations of the TTS model. 
    The speech is conditioned using the identity features extracted in the TFG model to align with the input identity.
}
    \label{fig:teaser}
\end{center}
}]
\nonumberfootnote{$^{*}$Equal contribution.}
\begin{abstract}
\vspace{-10pt}
The goal of this work is to simultaneously generate natural talking faces and speech outputs from text. We achieve this by integrating Talking Face Generation (TFG) and Text-to-Speech (TTS) systems into a unified framework.
We address the main challenges of each task: (1) generating a range of head poses representative of real-world scenarios, and (2) ensuring voice consistency despite variations in facial motion for the same identity.
To tackle these issues, we introduce a motion sampler based on conditional flow matching, which is capable of high-quality motion code generation in an efficient way.
Moreover, we introduce a novel conditioning method for the TTS system, which utilises motion-removed features from the TFG model to yield uniform speech outputs.
Our extensive experiments demonstrate that our method effectively creates natural-looking talking faces and speech that accurately match the input text. 
To our knowledge, this is the first effort to build a multimodal synthesis system that can generalise to unseen identities.
\end{abstract}    
\begin{textblock*}{.8\textwidth}[.5,0](0.5\textwidth, -0.98\textwidth)
\centering
{\small Project page with demo: \url{https://mm.kaist.ac.kr/projects/faces-that-speak}}
\end{textblock*}
\section{Introduction}
\label{sec:intro}
In recent years, the field of talking face synthesis has attracted growing interest, driven by the advancements in deep learning techniques and the development of services within the metaverse.
This versatile technology has diverse applications in movie and TV production, virtual assistants, video conferencing, and dubbing, with the goal of creating animated faces that are synchronised with audio to enable natural and immersive human-machine interactions.

Previous studies in deep learning-based talking face synthesis have focused on enhancing the controllability of facial movements and achieving precise lip synchronisation. Some notable works~\cite{das2020speech, song2022everybody, thies2020neural, yi2020audio, deng2019accurate, jiang2019disentangled, bulat2017far, ma2023styletalk} incorporate 2D or 3D structural information to improve motion representations. 
From this, recent research has naturally diverged into two primary strands along the target applications of TFG: one strand~\cite{zhou2020makelttalk, min2022styletalker, zhang2023sadtalker, wang2022one} concentrates on generating expressive facial movements only from audio conditions. Meanwhile, the other strand~\cite{zhou2021pose, liang2022expressive, burkov2020neural, wang2022progressive, hwang2023discohead, jang2023s} aims to enhance the controllability of talking faces by introducing a target video as an additional condition. Despite these advancements, the audio-driven TFG methods exhibit limitations, especially in scenarios like video production and AI chatbots, where video and speech must be generated simultaneously.

An emerging area of research is text-driven TFG, which is relatively under-explored compared to audio-driven TFG. 
Several studies~\cite{zhang2022text2video, wang2023text, ye2023ada} have attempted to merge TTS systems with TFG using a cascade approach, but suffered from issues like error accumulation or computational bottleneck. 
A very recent work~\cite{mitsui2023uniflg} uses latent features from TTS systems for face keypoint generation, yet still requires an additional stage for RGB video production. 
It highlights the challenges and complexities in integrating TFG and TTS systems into a cohesive and unified framework.

In this paper, we propose a unified framework, named \textbf{T}ext-\textbf{t}o-\textbf{S}peaking~\textbf{F}ace (\textbf{TTSF}), which integrates text-driven TFG and face-stylised TTS. The key to our method lies in analysing mutually complementary elements across distinct tasks and leveraging this analysis to construct an improved framework.
As illustrated in~\Fref{fig:teaser}, our framework is capable of simultaneously generating talking face videos and natural speeches given text and a face portrait.
To combine the different tasks in a single model, we tackle the primary challenges inherent in each task, TFG and TTS. 

Firstly, our approach enables the generation of a range of head poses that reflect real-world scenarios. To encompass dynamic and authentic facial movements, we propose a motion sampler based on Optimal-Transport Conditional Flow Matching (OT-CFM). This approach learns Ordinary Differential Equations (ODEs) to extract precise motion codes from a sophisticated distribution. Nonetheless, considerations need to be taken into account to apply OT-CFM to the motion sampling process.
Direct prediction of target motion by OT-CFM results in the generation of unsteady facial motions. To address this issue, we employ an auto-encoder-based noise reducer to mitigate feature noise through compression and reconstruction of latent features. The compressed features serve as the target motions for our motion sampler. This demonstrates an enhanced quality of the generated motion, particularly in terms of temporal consistency.

Secondly, we focus on the challenge of producing consistent voices, specifically when the input identity remains the same but facial motions differ. 
This problem arises from a fundamental inquiry in face-stylised TTS: How can we extract more refined speaker representations, influencing prosody, timbre, and accent, from a portrait image? We observe that facial motion in the source image affects the ability to identify the characteristics of the target voice. Nevertheless, this issue has been overlooked in all previous works~\cite{goto2020face2speech,wang2022residual,lee2023imaginary}, as they commonly omit a facial motion disentanglement module, a crucial component in the TFG system. With the benefit of integrating the TFG and TTS models into a system, we present a straightforward yet effective approach to condition the face-stylised TTS model. By eliminating motion features from the input portrait, our framework can generate speeches with the consistency of speaker identity.

In addition to the previously mentioned advantages of our framework, there are further benefits compared to cascade text-driven TFG systems: (1) our framework does not require an additional audio encoder, as it can be substituted with the text encoder in our system, and (2) the joint training eliminates the need for the fine-tuning process and yields better-synchronised lip motions in the generated outcomes.

Our contributions can be summarised as follows:
\begin{itemize}
    \item To our best knowledge, we are the first to propose a unified text-driven multimodal synthesis system with robust generalisation to unseen identities.
    \item We design the motion sampler based on OT-CFM that is combined with the auto-encoder-based noise reducer, by considering the characteristics of motion features.
    \item Our method preserves crucial speaker characteristics such as prosody, timbre, and accent by removing the motion factors in the source image, 
    \item With the comprehensive experiments, we demonstrate the proposed method surpasses the cascade-based talking face generation methods while producing speeches from the given text.
\end{itemize}

\begin{figure*}[!t]
    \centering
    \includegraphics[width=0.9\linewidth]{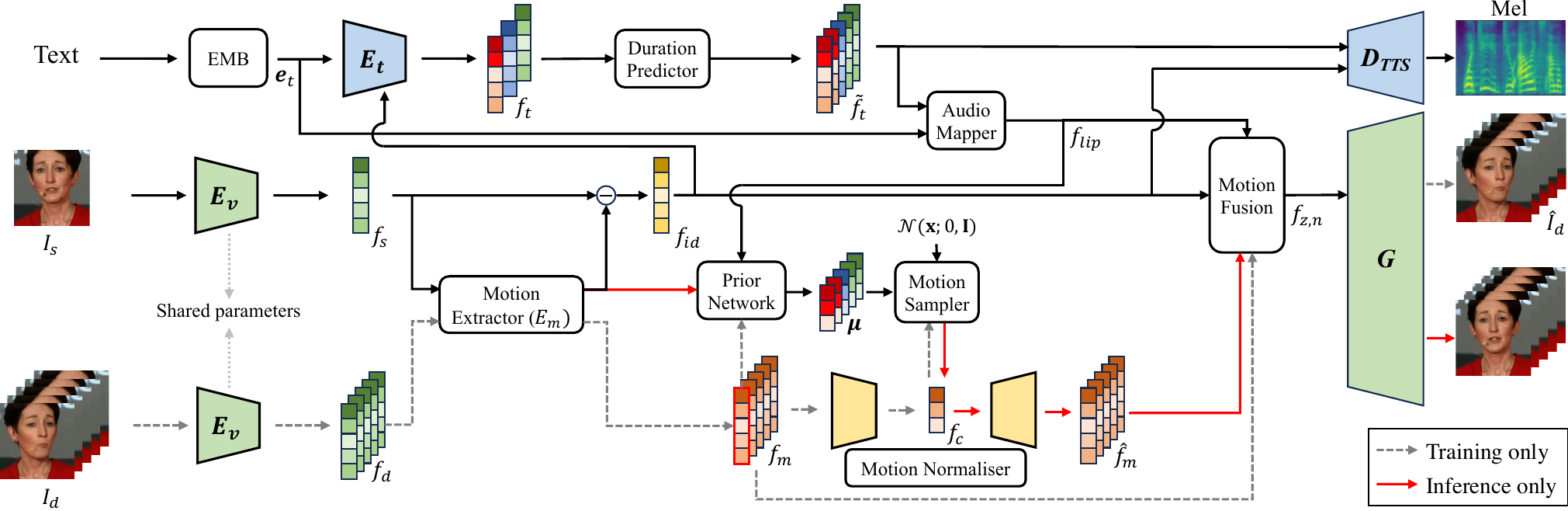}
    \vspace{-3mm}
    \caption{Overall architecture of our framework. The TTS model receives identity representations from the TFG model, while the TFG model takes conditions for natural motion generation from the TTS model. These complementary elements enhance our model's capabilities in generating both speech and talking faces. The EMB block denotes an embedding operation. The grey dashed arrow represents a path used only during the training process, and the red arrows represent paths used only during the inference process.}
    \vspace{-3mm}
    \label{fig:overall}
\end{figure*}
\section{Related Works}
\newpara{Audio-driven Talking Face Generation.}
Audio-driven Talking Face Generation (TFG) technology has captured considerable attention in the fields of computer vision and graphics due to its broad range of applications~\cite{chen2020comprises,zhu2021deep}. 
In the early works~\cite{fan2015photo,fan2016deep}, the focus is on situations with individual speakers, where a single model generates various talking faces based on a single identity. Recently, advancements in deep learning have facilitated the creation of more versatile TFG models~\cite{chung2017you,chen2019hierarchical,song2018talking,kr2019towards,zhou2019talking,prajwal2020lip,park2022synctalkface}. These models can generate talking faces by incorporating identity conditions as input. 
However, these studies overlook head movements, grappling with the difficulty of disentangling head poses from facial characteristics linked to identity.
To enhance natural facial movements, some studies integrate landmarks and mesh~\cite{das2020speech, song2022everybody, thies2020neural, yi2020audio} or leverage 3D information~\cite{deng2019accurate, jiang2019disentangled, bulat2017far, ma2023styletalk}. Despite these efforts, performance degradation occurs, especially in wild scenarios with low landmark accuracy.
Recent research branches~\cite{zhou2020makelttalk, min2022styletalker, zhang2023sadtalker, wang2022one} focus on generating vivid facial movements only from audio conditions. Another branch~\cite{zhou2021pose, liang2022expressive, burkov2020neural, wang2022progressive, hwang2023discohead, jang2023s} demonstrates improved controllability by introducing a target video as an additional condition. These studies showcase the creation of realistic talking faces with various facial movements, encompassing head, eyes, and lip movements.
However, these approaches rely on audio sources for TFG, limiting their applicability in multimedia scenarios lacking an audio source.

\newpara{Text-driven Talking Face Generation.}
Text-driven TFG is relatively less explored compared to the field of audio-driven TFG. 
Most previous works~\cite{kr2019towards,kumar2017obamanet,liu2022parallel, fried2019text,taylor2012dynamic} primarily focus on generating lip regions for text-based redubbing or video-based translation tasks. Recent works~\cite{zhang2022text2video,wang2023text,ye2023ada} have tried to incorporate Text-to-Speech (TTS) technology into the process of TFG through a cascade method. However, it's worth noting that the cascade method encounters bottlenecks in terms of both performance and inference time~\cite{choi2023reprogramming}. To tackle this issue, the latest study~\cite{mitsui2023uniflg} has delved into the latent features of TTS to generate keypoints for talking faces. This exploration provides evidence that leveraging the latent features of a TTS model is advantageous in substituting the latent of an audio encoder for TFG.

In this paper, we unify TTS and TFG tasks to generate speech and talking face videos concurrently. Furthermore, we extend the application of TTS in TFG by conditioning the target voice with the input identity image. As a result, our model can generate a diverse range of talking face videos using only a static face image and text as input.

\newpara{Text-to-Speech.}
Text-to-Speech (TTS) systems aim to generate natural speech from text inputs, evolving from early approaches to recent end-to-end methods~\cite{lee2021multi, black2007statistical, shen2018natural, ren2019fastspeech, kim2020glow, popov2021grad, mehta2023matcha}. 
Despite their success, unseen-speaker TTS systems face a challenge in requiring substantial enrollment data for accurate voice reproduction. While prior works~\cite{jia2018transfer, min2021meta, chen2021adaspeech, lee2022pvae, huang2022generspeech} extract speaker representations from speech data, obtaining sufficient high-quality utterances is challenging. 
Recent studies have incorporated face images for speaker representation~\cite{goto2020face2speech, wang2022residual, lee2023imaginary}, aiming to capture correlations between visual and audio features. However, these models often neglect motion-related factors in face images, leading to challenges in generating consistent desired voices when the input identity remains constant but the motion varies.

In this paper, to tackle this issue, we leverage the motion extractor of TFG to eliminate the motion features from the source image. The motion-normalised feature is then fed into the TTS system as a conditioning factor, aiding the TTS model in producing consistent voices.


\section{Method}
In~\Fref{fig:overall}, we propose a unified architecture, named \textbf{TTSF}, which integrates TFG and TTS pipelines.
In the TTS model, the text input is embedded as $\boldsymbol{e}_{t}$ by an embedding layer. The text encoder $E_t$ maps this embedding to the text feature $f_{t} \in \mathbb{R}^{l_t \times d}$, where $l_t$ and $d$ denote the token length and hidden dimension, respectively. The duration predictor then upsamples $f_{t}$ to $\Tilde{f}_{t} \in \mathbb{R}^{l_m \times d}$ to align with the target mel-spectrogram's length $l_m$. The $\Tilde{f}_{t}$ is subsequently passed into the TTS decoder $D_{TTS}$ to predict the target mel-spectrogram. Both $E_t$ and $D_{TTS}$ are conditioned with the identity feature ${f}_{id}$ from the TFG model to incorporate the characteristics of the target speaker.
In the TFG model, the source image $I_{s}$ and driving frames $I_{d} \in \mathbb{R}^{t \times c \times h \times w}$ pass through the shared visual encoder $E_{v}$, yielding visual features ${f}_s$ and ${f}_d$ for the source and target, respectively. The motion extractor encodes motion features from the input, obtaining the identity feature ${f}_{id}$ by subtracting the motion feature from ${f}_s$. The target motion feature is denoted as $f_{m}$. With the motion fusion module, $f_{id}$, $f_{m}$, and the audio mapper output $f_{lip}$ are aggregated and then, input into the TFG generator $G$ to generate videos $\hat{I}_d$ with desired motions.
To produce variational facial movements during inference, we propose a conditional flow matching-based motion sampler. Additionally, we introduce an auto-encoder-based motion normaliser aimed at reducing the noise in the sampled motions. The feature $f_c$, compressed by the normaliser, serves as the motion sampler's target during training. Consequently, our framework synthesises natural talking faces and speeches from a single portrait image and text condition.

\subsection{Baseline for Talking Face Generation}
\label{subsec:tfg}
\newpara{Motion Extractor.}
Previous research in the fields of motion transfer~\cite{siarohin2019first, siarohin2021motion, wang2022latent} and TFG~\cite{wang2021one, jang2023s} has identified the presence of a reference space that only contains individual identities. Formally, we can express this as $E_{v}(I) = f_{id} + f_{m}$, where $I$ is the input image, $E_{v}$ is the visual encoder, $f_{id}$ is an identity feature, and $f_{m}$ is a motion feature. In our framework, the motion extractor $E_{m}$ learns the subtraction of identity feature $f_{id}$ from the visual feature:
$E_{m}(E_{v}(I)) = f_{m} = f - f_{id}.$
Our motion extractor follows the architecture of LIA~\cite{wang2022latent}, featuring a 5-layer MLP and trainable motion codes under an orthogonality constraint. This constraint facilitates the representation of diverse motions with compact channel sizes. Unlike LIA, which computes relative motion between source and target images, our motion extractor independently extracts identity and motion features. This distinction is crucial for integrating TFG and TTS models, where the identity feature conditions TTS to generate consistent voice styles robust to facial motions.

\newpara{Motion Fusion and Generator.}
To establish a baseline for generating both talking faces and speeches, we consider two key aspects in designing the TFG generator $G$: (1) memory efficiency and (2) resilience to unseen identity generation. To reflect these, we avoid using an inversion network, known for its computational heaviness, and opt for a flow-based generator that focuses on learning coordinate mapping. For our generator, we choose LIA's one, which employs a StyleGAN~\cite{abdal2020image2stylegan++}-styled generator as a baseline.

However, LIA is explicitly tailored for face-to-face motion transfer and does not account for generating lip movements synchronised with an audio source. To apply LIA to TFG, specific considerations are needed. 
In the training process, the lack of augmentation for target frames leads to the model replicating lip motions from the target frames rather than from audio sources. In response to this, inspired by FC-TFG~\cite{jang2023s}, we regulate lip motions by incorporating audio features into specific $n$-th layers of the decoder. The fusion process involves a straightforward linear operation:
\begin{equation}
f_{z,n} =\left\{\begin{matrix}
 f_{id} + f_{m} &i \in \{\text{non-lip motion layers}\}\\ 
 f_{id} + f_{lip} &i \in \{\text{lip motion layers}\},
\end{matrix}\right.
\end{equation}
where, $f_{m}$ denotes the target motions extracted from target frames and $f_{lip}$ denotes the output of the audio mapper, representing lip motion features.
In the end, we generate the final videos $\hat{I}_d$ by inputting the style feature $f_{z,n}$ into the TFG generator $G$.

\newpara{Audio Mapper.} 
\begin{figure}[!t]
    \centering
    \includegraphics[width=0.65\linewidth]{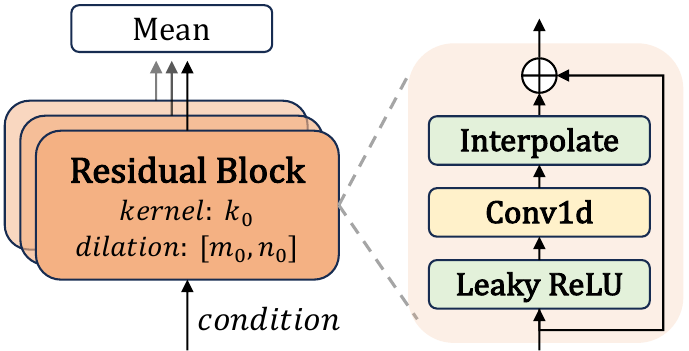}
    \vspace{-2mm}
    \caption{The architecture of the audio mapper. The \emph{condition} denotes the concatenated feature of text embedding $\boldsymbol{e}_{t}$, upsampled text feature $\Tilde{f}_t$, and energy, which is a norm of $\Tilde{f}_t$.
    \vspace{-3mm}
    }
    \label{fig:audiomap}
\end{figure}
Unlike the cascade text-driven TFG, our framework does not require extracting acoustic features using an audio encoder.
Instead, we utilise the intermediate representations of the TTS system, serving a definite  purpose: Generating natural lip motion with the TFG generator.
This feature is crafted by aggregating the concatenated features of text embedding $\boldsymbol{e}_t$, the upsampled text feature $\Tilde{f}_t$, and energy which is an average from $\Tilde{f}_t$ along the channel axis.
The text embedding enables the TFG model to grasp phoneme-level lip representation, while the upsampled text feature and energy contribute to capturing intricate lip shapes aligned with the generated speech sound.
To aggregate these different types of features, we use Multi-Receptive field Fusion (MRF) module~\cite{kong2020hifi}. As illustrated in~\Fref{fig:audiomap}, the MRF module comprises multiple residual blocks, each characterised by 1D convolutions with distinct kernel sizes and dilations. 
This diverse configuration enables the module to observe both fine and coarse details in the input along the time axis.
To avoid potential artifacts at the boundaries of motion features caused by temporal padding operations, we intentionally remove the padding operation and introduce temporal interpolation.
Consequently, our framework achieves well-synchronised lip movements while effectively capturing the characteristics of the generated speech.

\newpara{Training Objectives.}
We use a non-saturating loss~\cite{goodfellow2020generative} in adversarial training:
{\small
\begin{equation}
\begin{split}
    \mathcal{L}_{GAN} = \min_{G}\max_{D}\Bigl(&\mathbb{E}_{I_d}[\log(D(I_d))] \\
    &+ \mathbb{E}_{f_{z,n}}[\log(1-D(G({f_{z,n}}))] \Bigr).
\end{split}
\end{equation}
}
For pixel-level supervision, we use $L1$ reconstruction loss and Learned Perceptual Image Patch Similarity (LPIPS) loss~\cite{zhang2018unreasonable}. The reconstruction loss $\mathcal{L}_{rec}$ is formulated as:
{\small
\begin{equation}
    \mathcal{L}_{rec} = \parallel \hat{I}_d - I_{d} \parallel_1 + \frac{1}{N_f}\sum_{i=1}^{N_f} \parallel \phi(\hat{I}_d)_i - \phi(I_{d})_i \parallel_2,
\end{equation}
}
where $\phi$ is a pretrained VGG19~\cite{simonyan2014very} network, and $N_f$ is the number of feature maps.
To preserve facial identity after motion transformation, we apply an identity-based similarity loss~\cite{richardson2021encoding} using a pretrained face recognition network 
$\mathcal{L}_{id} = 1 - \cos \left( E_{id}\left(\hat{I}_{d}\right), E_{id}\left(I_{d}\right) \right).$
Finally, To generate well-synchronised videos according to the input audio conditions, We use the modified SyncNet introduced in~\cite{jang2023s} to enhance our model's lip representations.
We minimise the following sync loss: $\mathcal{L}_{s y n c}=1-\cos \left(S_v\left(\hat{I}_{d}\right), S_a\left(A_s\right)\right),$
where $S_a$, $S_v$, and $A_s$ denote the audio encoder, video encoder of SyncNet, and input audio source.

\subsection{Variational Motion Sampling}
\newpara{Preliminary: Conditional Flow Matching.}
In this subsection, we present an outline of Optimal-Transport Conditional Flow Matching (OT-CFM). Our exposition primary adheres to the notation and definitions in~\cite{lipman2022flow,mehta2023matcha}.

Let $\x \in \real^d$ be the data sample from the target distribution $q(\x)$, and $p_0(\x)$ be tractable prior distribution. Flow matching generative models aim to map $\x_0\sim p_0(\x)$ to $\x_1$ by constructing a probability density path $p_t: [0,1]\times \real^d \rightarrow \real_{>0}$, such that $p_1(\x)$ approximates $q(\x)$. Consider an arbitrary Ordinary Differential Equation (ODE): 
\begin{align}
\frac{d}{dt}\phi_t(\x)
& = \boldsymbol{v}_t(\phi_t(\x))
\text{,}
\qquad
\phi_0(\x)
= \x
\text{,}
\label{eq:ode}
\end{align}
where the vector field $\boldsymbol{v}_t: [0,1] \times \real^d \rightarrow \real^d$ generates the flow $\phi_t: [0, 1] \times \real^d \rightarrow \real^d$. This ODE is associated with $p_t$, and it is sufficient to produce realistic data if a neural network can predict an accurate vector field $\boldsymbol{v}_t$.

Suppose there exists the optimal vector field $\boldsymbol{u}_t$ that can generate accurate ${p}_t$, then the neural network $\boldsymbol{v}_t(\x;\theta)$ can be trained to estimate the vector field $\boldsymbol{u}_t$. However, in practice, it is non-trivial to find the optimal vector field $\boldsymbol{u}_t$ and the target probability ${p}_t$. To address this,~\cite{lipman2022flow} leverages the fact that estimation of conditional vector field is equivalent to estimation of the unconditional one, i.e., 
\begin{equation}
\begin{split}
    &\min_{\theta}\mathbb{E}_{t,p_t(\x)}\Vert  \boldsymbol{u}_t(\x) - \boldsymbol{v}_t(\x; \theta)\Vert^2 \\
    & \equiv \min_{\theta}\mathbb{E}_{t,q(\x_1),p_t(\x|\x_1)}\Vert \boldsymbol{u}_t(\x\vert \x_1) - \boldsymbol{v}_t(\x; \theta)\Vert^2 
\end{split}
\end{equation}
with boundary condition $p_0(\x|\x_1)=p_0(\x)$ and $p_1(\x|\x_1)=\mathcal{N}(\x|\x_1, \sigma^2 \boldsymbol{I})$ for sufficiently small $\sigma$. 

Meanwhile, \cite{lipman2022flow} further generalise this technique with noise condition $\x_0 \sim \mathcal{N}(0,1)$, and define OT-CFM loss as:

\begin{equation}
\begin{split}
\mathcal{L}_{\mathrm{OT-CFM}} (\theta) =\mathbb{E}&_{t, q(\x_1), p_0(\x_0)}\Vert\boldsymbol{u}^{\mathrm{OT}}_t(\phi^{\mathrm{OT}}_t(\x_0)| \x_1)\\
&-\boldsymbol{v}_t(\phi^{\mathrm{OT}}_t(\x_0) | \boldsymbol{\mu}; \theta) \Vert^2,
\end{split}
\end{equation}
where $\boldsymbol{\mu}$ is the predicted frame-wise mean of $\x_1$ and $\phi^{\mathrm{OT}}_{t}(\x_0) = (1- (1-\sigma_{\mathrm{min}})t)\x_0 + t \x_1$ is the flow from $\x_0$ to $\x_1$. The target conditional vector field become $\boldsymbol{u}^{\mathrm{OT}}_t(\phi^{\mathrm{OT}}_t(\x_0)\vert \x_1) = \x_1-(1-\sigma_{\mathrm{min}})\x_0$, which enables the improved performance with its inherent linearity. In our work, we use fixed value of $\sigma_{min}=10^{-4}$.

\newpara{Prior Network.} 
The prior serves as the initial condition for OT-CFM, facilitating the identification of the optimal path to $x_1$. 
During training, our prior network takes the first motion $f_{m,0}$ of target motion sequence $f_{m}$ and the acoustic feature $f_{lip}$ as inputs. 
We structure the prior network with a 4-layer conformer~\cite{gulati2020conformer}, where the input is formed by the summation of $f_{m,0}$ and $f_{lip}$.
Note that the first motion is replaced as the source image's motion in inference.

\newpara{OT-CFM Motion Sampler.}
The objective of our motion sampler is to sample a sequence of natural motion codes from the prior $\boldsymbol{\mu}$. 
During training, this module aims to predict target motions $f_{m}$. However, in our experiments, we observed that directly regressing $f_{m}$ (equivalent to setting $\x_1$ as $f_{m}$) leads to producing shaky motions during inference. We expect that this is due to the characteristics of the StyleGAN-styled decoder. Each channel of the decoder plays a semantically meaningful role in generating detailed facial attributes. Therefore, when the motion sampler fails to successfully estimate the vector field, it directly impacts the final outcomes.
To address this issue, we introduce an auto-encoder-based motion normaliser that compresses feature and reconstructs them into the target motion $f_{m}$. 
The compressed motion features $f_{c}$ serve as $\x_1$ in OT-CFM. 

\newpara{Training Objectives.}
The reconstruction loss for training our motion normaliser is defined as Mean Square Error (MSE) loss between the target motion $f_{m}$ and the reconstructed motion $\hat{f}_{m}$ as follows: $\mathcal{L}_{AE} = \parallel \hat{f}_{m} - f_{m} \parallel_2.$
Moreover, as motion decoding commences from random noise $\mathcal{N}(\boldsymbol{\mu},I)$ at inference, our objective is to minimise the distance between the prior $\boldsymbol{\mu}$ and compressed target motion $f_{c}$.
Considering the output of prior network $\boldsymbol{\mu}$ as parameterising the input noise for the decoder, it is natural to view the encoder output ${\boldsymbol{\mu}}$ as a normal distribution $\mathcal{N}({\boldsymbol{\mu}},I)$. Following~\cite{popov2021grad}, we compute a negative log-likelihood prior loss:
\begin{equation}
\label{eq:loss_enc}
    \mathcal{L}_{prior} = -\sum_{j=1}^{T}{\log{\varphi(f_{c,j};{\mu}_{j}, I)}},
\end{equation}
where $\varphi(\cdot;{\mu}_{i},I)$ represents the probability density function of $\mathcal{N}({\mu}_{i}, I)$, and $T$ denotes the temporal length of motions.

\begin{figure*}[!t]
    \centering
    \includegraphics[width=0.9\linewidth]{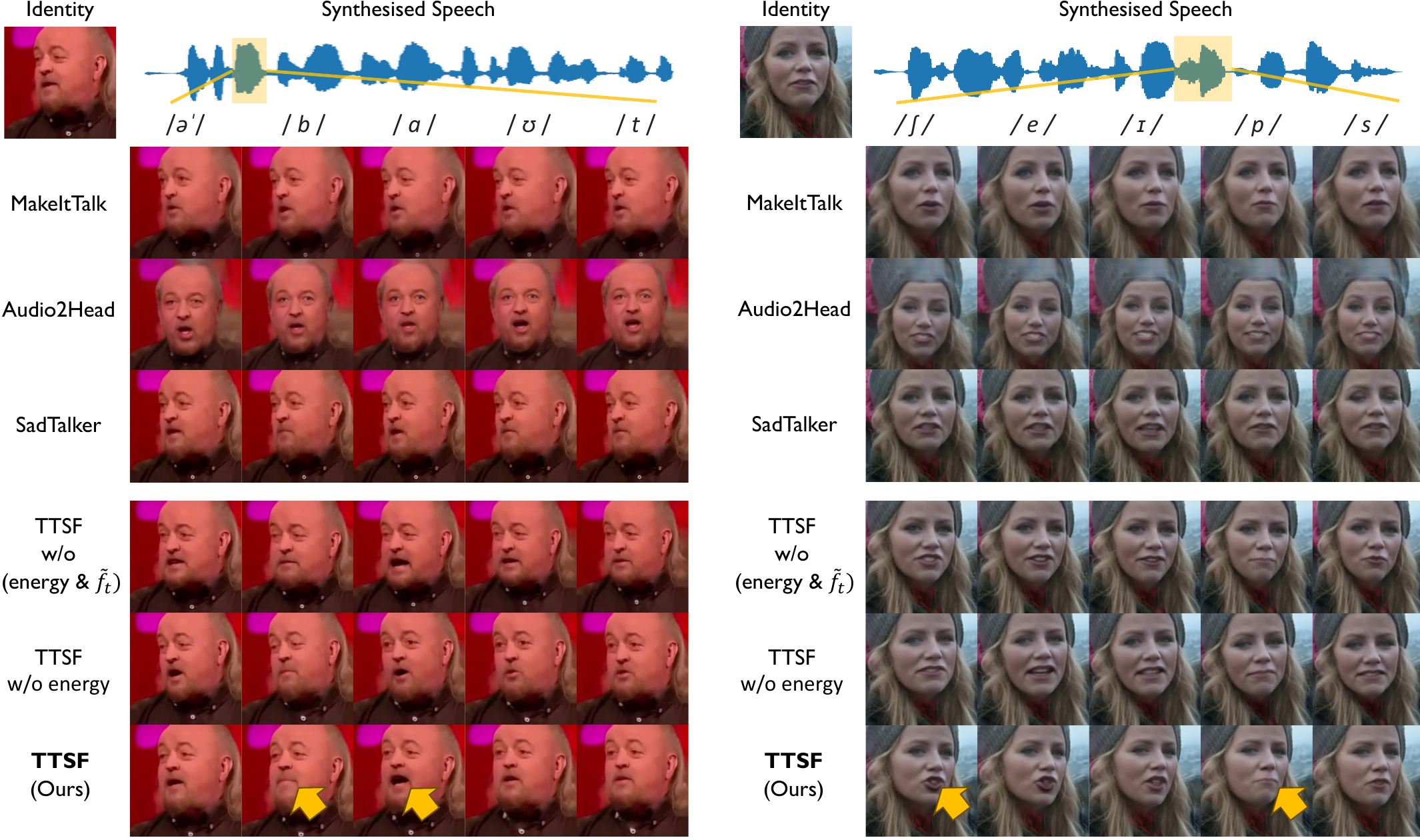}
    \vspace{-2mm}
    \caption{Qualitative Results. We compare our method with several baselines listed in~\Cref{tab:results_lrs2}. 
    Our approach outperforms all the baselines in terms of generating natural facial motions, encompassing lip shape and head pose. 
    MakeItTalk and SadTalker exhibit smaller variance in head poses, while Audio2Head fails to preserve the source identity. We emphasis that our TTSF system can generate sophisticated lip shapes, reflecting both linguistic and acoustic information from our TTS model. 
    \vspace{-3mm}
    }
    \label{fig:sota}
\end{figure*}

\subsection{Text-to-Speech Synthesis}
Our TTS system aims to produce well-stylised speech from a single portrait, acquired in in-the-wild setting.
In this context, we define the in-the-wild environment as follows: (1) The model is exposed to previously unseen facial data, and (2) the facial images exhibit various facial poses.
First, since we cannot access to the identity labels to unseen speakers, we condition our model with image embedding. Second, our emphasis is on the advantages of our framework. 
By integrating TFG and TTS systems, we can utilise the identity feature $f_{id}$, a motion-removed feature, obtained from the TFG model. Consequently, our TTS model is capable of generating speeches robust to various facial motions in image, maintaining consistency in the style of voice.

Our system is based on Matcha-TTS~\cite{mehta2023matcha}, an OT-CFM-based TTS model known for synthesising high-quality speeches in a few synthesis steps. We input the identity feature $f_{id}$ to both encoder and decoder. With this minimal variation, our model is trained with prior, duration, and OT-CFM losses, as outlined in~\cite{mehta2023matcha}. These losses are collectively denoted as $L_{TTS}$. Finally, we convert the generated mel-spectrogram by using a pretrained vocoder~\cite{kong2020hifi}.



\newpara{Final Loss.} The final loss is calculated as the sum of the aforementioned losses, represented as follows:
\begin{equation}
\begin{split}
    \mathcal{L}_{total} = \lambda_{1}\mathcal{L}_{GAN} + 
    \lambda_{2}\mathcal{L}_{rec} + \lambda_{3}\mathcal{L}_{id} + \lambda_{4}\mathcal{L}_{sync} \\ + \lambda_{5}\mathcal{L}_{OT-CFM} + \lambda_{6}\mathcal{L}_{ae} + \lambda_{7}\mathcal{L}_{prior} + \lambda_{8}\mathcal{L}_{tts},
\end{split}
\end{equation}
where hyperparameters $\lambda$ are introduced to balance the scale of each loss. Each $\lambda$ controls the relative importance of its corresponding loss term. Empirically, the values of $\lambda_{1}$, $\lambda_{2}$, $\lambda_{3}$, $\lambda_{4}$, $\lambda_{5}$, $\lambda_{6}$, $\lambda_{7}$, and $\lambda_{8}$ are set to 0.1, 1, 0.3, 0.1, 0.1, 1, 0.1, and 1.

\section{Experiments}

\subsection{Experimental Setup}

\newpara{Dataset.}
Our framework is trained on the LRS3~\cite{afouras2018lrs3} dataset, which consists of both talking face videos and transcription labels.
LRS3 consists of videos captured during indoor shows of TED or TEDx.
We evaluate our model on VoxCeleb2~\cite{chung2018voxceleb2} and LRS2~\cite{Afouras18c} datasets, which contain more challenging examples than LRS3 since many videos are shot outdoors. We randomly select a subset of videos from each dataset to evaluate the performance of our framework.

\newpara{Implementation Details.}
First of all, we pretrain a Matcha-TTS~\cite{mehta2023matcha} model on the LRS3 dataset for 2,000 epochs and then jointly train with the talking face generation model for 40 epochs. Our focus is on the manipulation of seven specific layers within the generator, namely layers 1 to 7. Furthermore, we exclusively input the audio feature into two specific layers, namely layers 6 and 7. 
Our motion sampler is trained with 32-frame videos and then inferences with all frames of each video. 
Audio data is sampled to 16kHz, and converted to mel-spectrogram with a window size of 640, a hop length of 160, and 80 mel bins.
To update our model, we employ the Adam optimiser~\cite{kingma2014adam} with a learning rate set at ${1e{-4}}$. The entire framework is implemented using PyTorch~\cite{paszke2019pytorch} and is trained across eight 48GB A6000 GPUs.

\begin{table}[!t]
    \centering
    \resizebox{0.95\linewidth}{!}
    {
        \renewcommand{\arraystretch}{1.3}
        \begin{tabular}{l|c|cc|c|c}
        \hline\hline
        \multirow{2}{*}{\bf Models} & \multirow{2}{*}{\bf Audio} & \multicolumn{2}{|c}{\bf Video Quality} & \multicolumn{1}{|c}{\bf Synchronisation} & \multicolumn{1}{|c}{\bf Diversity} \\ 
        \cline{3-6}
         & & {FID}$\downarrow$ & {ID-SIM}$\uparrow$\quad & {LSE-C}$\uparrow$\quad & {DIV}$\uparrow$\quad \\ 
        \hline
        Ground Truth & - & 0.00 & 1.00 & 5.360 & 0.168 \\
        \hline
        MakeItTalk~\cite{zhou2020makelttalk} & GT & 24.213 & 0.845 & 2.674 & 0.102 \\
        MakeItTalk & TTS & 25.168 & 0.850 & 3.487 & 0.095 \\
        Audio2Head~\cite{wang2021audio2head} & GT & 41.721 & 0.217 & 4.607 & \bf 0.149 \\
        Audio2Head & TTS & 42.262 & 0.225 & 5.478 &  0.145 \\
        SadTalker~\cite{zhang2023sadtalker} & GT & 20.771 & 0.854 & 4.978 & 0.109 \\
        SadTalker & TTS & 20.729 & 0.859 & \bf 6.256 & 0.111 \\
        \hline
        \bf TTSF (Ours) & - & \bf 18.348 & \bf 0.864 & 5.686 & 0.143  \\
        \hline\hline
        \end{tabular}
    }
    \vspace{-1mm}
\caption{Comparison with the state-of-the-art methods on LRS2 in the one-shot setting. The \textbf{Audio} column refers to the speech source for generation (GT: ground truth, TTS: synthesised audio.)}
\vspace{-3mm}
\label{tab:results_lrs2}
\end{table}

\newpara{Evaluation Metrics.} In our quantitative assessments for TFG, we employ a range of evaluation metrics introduced in previous works. To assess the visual quality of the generated videos, we employ the Fréchet Inception Distance (FID) score and ID Similarity (ID-SIM) score using a pretrained face recognition model~\cite{huang2020curricularface}. To measure the accuracy of mouth shapes and lip sync, we utilise the Lip Sync Error Confidence (LSE-C), a metric introduced in~\cite{chung2017out}. For the diversity of the generated head motions, we calculate the standard deviation of the head motion feature embeddings extracted from the generated frames using Hopenet~\cite{Ruiz_2018_CVPR_Workshops}, following the approach introduced in~\cite{zhang2023sadtalker}.

For the evaluation of TTS performance, we compute Word Error Rate (WER), Mel Cepstral Distortion (MCD), the cosine similarity (C-SIM) between x-vectors~\cite{snyder2018x} of the target and synthesised speech, as well as the Root Mean Square Error (RMSE) for F0. 
WER and MCD represent the intelligibility and naturalness of speech, respectively. 
C-SIM and RMSE measure the voice similarity to the target speaker. For WER, we use a publicly available speech recognition model of \cite{radford2023robust}. 

\begin{table}[!t]
    \centering
    \resizebox{0.95\linewidth}{!}
    {
        \renewcommand{\arraystretch}{1.3}
        \begin{tabular}{l|c|c|c}
        \hline\hline
        \multirow{2}{*}{\textbf{Models}} & \bf Video Quality & \bf Synchronisation & \bf Diversity \\ 
        \cline{2-4}
         &  {ID-SIM}$\uparrow$\quad & {LSE-C}$\uparrow$\quad & {DIV}$\uparrow$\quad \\ 
        \hline
        MakeItTalk~\cite{zhou2020makelttalk} & 0.841 & 3.529 & 0.094 \\
        Audio2Head~\cite{wang2021audio2head} & 0.151 & 5.738 &  \textbf{0.146} \\
        SadTalker~\cite{zhang2023sadtalker} & 0.855 & \textbf{6.310} & 0.111 \\
        \hline
        \bf{TTSF} (Ours) & \textbf{0.876} & 5.721 & 0.143  \\
        \hline\hline
        \end{tabular}
    }
    \vspace{-1mm}
\caption{Comparison with the state-of-the-art methods on VoxCeleb2 in the one-shot setting. The previous audio-driven TFG models are cascaded with our TTS model to generate talking faces from text.}
\label{tab:results_voxceleb2}
\end{table}
\subsection{Comparison with State-of-the-art Methods}
\newpara{Text-driven Talking Face Generation.} 
We compare several state-of-the-art methods (MakeitTalk~\cite{zhou2020makelttalk}, Audio2Head~\cite{wang2021audio2head}, and SadTalker~\cite{zhang2023sadtalker}) for the text-driven talking head video generations by attaching our TTS model to the previous audio-driven TFG models in the cascade method.
To simulate a one-shot talking face generation scenario, we evaluate the baselines on the in-the-wild datasets, LRS2 and VoxCeleb2. 
As shown in~\Tref{tab:results_lrs2}, the proposed model outperforms every audio- and cascade text-driven method in terms of video quality (FID, ID-SIM) on LRS2. 
Additionally, we present experimental results on the VoxCeleb2 dataset in~\Tref{tab:results_voxceleb2}. Since this dataset does not contain text transcription according to the speech in the video, our framework generates both speech and a talking face by inputting a single frame from a VoxCeleb2 video and a randomly selected transcription from LRS3. 
Similar to the experimental results on LRS2, our framework exhibits superior performance in ID-SIM score.
On the other hand, the proposed model records a lower synchronisation score compared to SadTalker using the LSE-C metric. However, given that the LSE-C metric relies significantly on a pretrained model, a more useful evaluation of lip synchronisation can be achieved through perceptual judgement by humans, as assessed in user studies. The qualitative assessment in~\Cref{ex:user_studies} shows that our method produces perceptually better synchronised output compared to the baseline.
Although Audio2Head shows the best diversity score, it records the lowest scores in video quality metrics. We also observe that Audio2Head completely fails to generate a natural video when the input source image is not located in the centre of the screen. On the other hand, our proposed method achieves high scores in both video quality and diversity metrics.
Considering the aforementioned issues, our framework demonstrates robust generalisation to unseen data when conducting multimodal synthesis encompassing both video and speech.

\newpara{Face-stylised Text-to-Speech.}
To evaluate the generalisability of our TTS system, we compare our model to Face-TTS~\cite{lee2023imaginary}, which is a state-of-the-art method of face-stylised TTS.
For the evaluation, we simulate two scenarios on LRS2 dataset: (1) \textit{w/ motion}, where the TTS model is conditioned with source image embedding, i.e., $f_{id}+f_m$; (2) \textit{w/o motion}, where the model is conditioned with only identity feature $f_{id}$. 
The results are shown in Table~\ref{tab:results_lrs2}. 
While the proposed model shows slight deviance in MCD, it clearly outperforms the baseline in WER, C-SIM, and RMSE, demonstrating its superiority in intelligibility and voice similarity.
More importantly, when we consider motion features together as our speaker condition, the generation performance is significantly degraded, especially in voice similarity. 
It indicates the benefits of unifying TFG and TTS systems, highlighting the advantages of their integration.


\begin{table}[!t]
    \centering
    \resizebox{0.9\linewidth}{!}
    {
        \renewcommand{\arraystretch}{1.3}
        \begin{tabular}{l|c|c|cc}
        \hline\hline
        \multirow{2}{*}{\textbf{Models}} &\multicolumn{1}{c|}{\textbf{Intel.}} &\multicolumn{1}{c|}{\textbf{Nat.}} & \multicolumn{2}{c}{\textbf{Voice similarity}}\\ 
        \cline{2-5}
         & {WER}$\downarrow$\quad &{MCD}$\downarrow$\quad &{C-SIM}$\uparrow$\quad &{RMSE}$\downarrow$  \\ \hline
        Ground Truth &6.35 &-- &-- &-- \\ \hline
        Face-TTS~\cite{lee2023imaginary}  &18.02 & \textbf{6.85} & 0.272 &52.33 \\ 
        Ours (w/ motion) &15.68 & 7.43 & 0.451 &50.67  \\ 
        Ours (w/o motion) &\textbf{14.56} & 7.23 & \textbf{0.593} &\textbf{48.52}\\ 
        \hline\hline
        \end{tabular}
    }
\caption{Quantitative results of synthesised speech. Intel. and Nat. denote intelligibility and naturalness of audio, respectively.}
\label{tab:tts}
\end{table}

\begin{table}[t]
    \centering
    \resizebox{0.95\linewidth}{!}
    {
            \renewcommand{\arraystretch}{1.3}
            \begin{tabular}{l|c|c|c}
            \hline\hline
            \multirow{2}{*}{\textbf{Models}}  & \bf Lip Sync & \bf Motion  &  \bf Video  \\ 
            & \bf Accuracy &\bf Naturalness &\bf Realness \\
            \cline{1-4}
            MakeItTalk~\cite{zhou2020makelttalk} & 2.44 $\pm$ 0.07 & 2.79 $\pm$ 0.09 & 2.79 $\pm$ 0.08 \\
            Audio2Head~\cite{wang2021audio2head} & 2.73 $\pm$ 0.08 & 2.92 $\pm$ 0.09 & 2.48 $\pm$ 0.09 \\
            SadTalker~\cite{zhang2023sadtalker} & 2.78 $\pm$ 0.08  & 2.80 $\pm$ 0.09 & 2.90 $\pm$ 0.09 \\
            \hline
            \bf{TTSF(Ours)} & \bf 4.09 $\pm$ 0.07 & \bf 3.85 $\pm$ 0.07 & \bf 3.87 $\pm$ 0.07 \\
            \hline\hline
            \end{tabular}
    }
\caption{\small MOS evaluation results. MOS is presented with 95\% confidence intervals. Note that the previous audio-driven TFG models are cascaded with our TTS model.}
\vspace{-1mm}
\label{tab:userstudy}
\end{table}

\subsection{Qualitative Evaluation}
\label{ex:user_studies}
\newpara{User Study.} 
We evaluate the synthesised videos through a user study involving 40 participants, each providing opinions on 20 videos. Reference images and texts were randomly selected from the LRS2 test split to create videos using MakeItTalk~\cite{zhou2020makelttalk}, Audio2Head~\cite{wang2021audio2head}, SadTalker~\cite{zhang2023sadtalker}, and our proposed method. 
Mean Opinion Scores (MOS) are used for evaluation, following the approach in~\cite{liang2022expressive,zhou2021pose,jang2023s}. Participants rate each video on a scale from 1 to 5, considering lip sync quality, video realness, and head movement naturalness. 
The order of methods within each video clip is randomly shuffled. 
The results in~\Tref{tab:userstudy} indicate that our method outperforms existing methods in generating talking face videos with higher lip synchronisation and natural head movement.

\newpara{Analysis on Qualitative Results.}
We visually present our qualitative results in ~\Fref{fig:sota}. MakeItTalk fails to produce precise lip motions aligned with the synthesised speech, and Audio2Head struggles to preserve identity information. SadTalker can generate well-synchronised lip motions but is limited in facial movements. 
In contrast, our approach exhibits more dynamic facial movements and can generate vivid lip motions that reflect both linguistic and acoustic information. 
For instance, it can be seen that our model's lip motions are precisely aligned to the pronunciation of the speeches (refer to the yellow arrows).
The accuracy and the details demonstrate that our method can generate realistic and expressive talking faces.

\begin{figure}[!t]
    \centering
    \subfloat[][Face-TTS~\cite{lee2023imaginary}]{
        \includegraphics[width=0.45\linewidth]{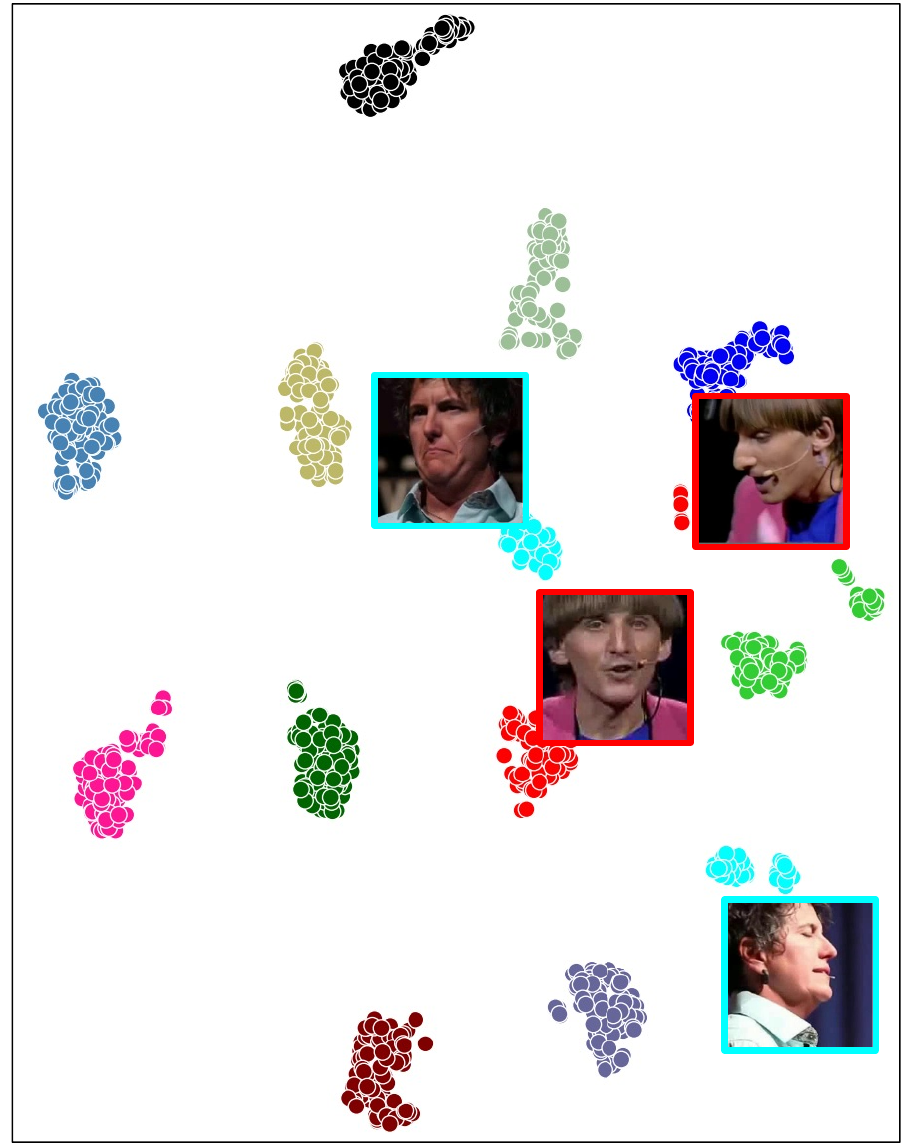}
        \label{fig:face-tts}
    }
    \subfloat[][Ours]{
        \includegraphics[width=0.45\linewidth]{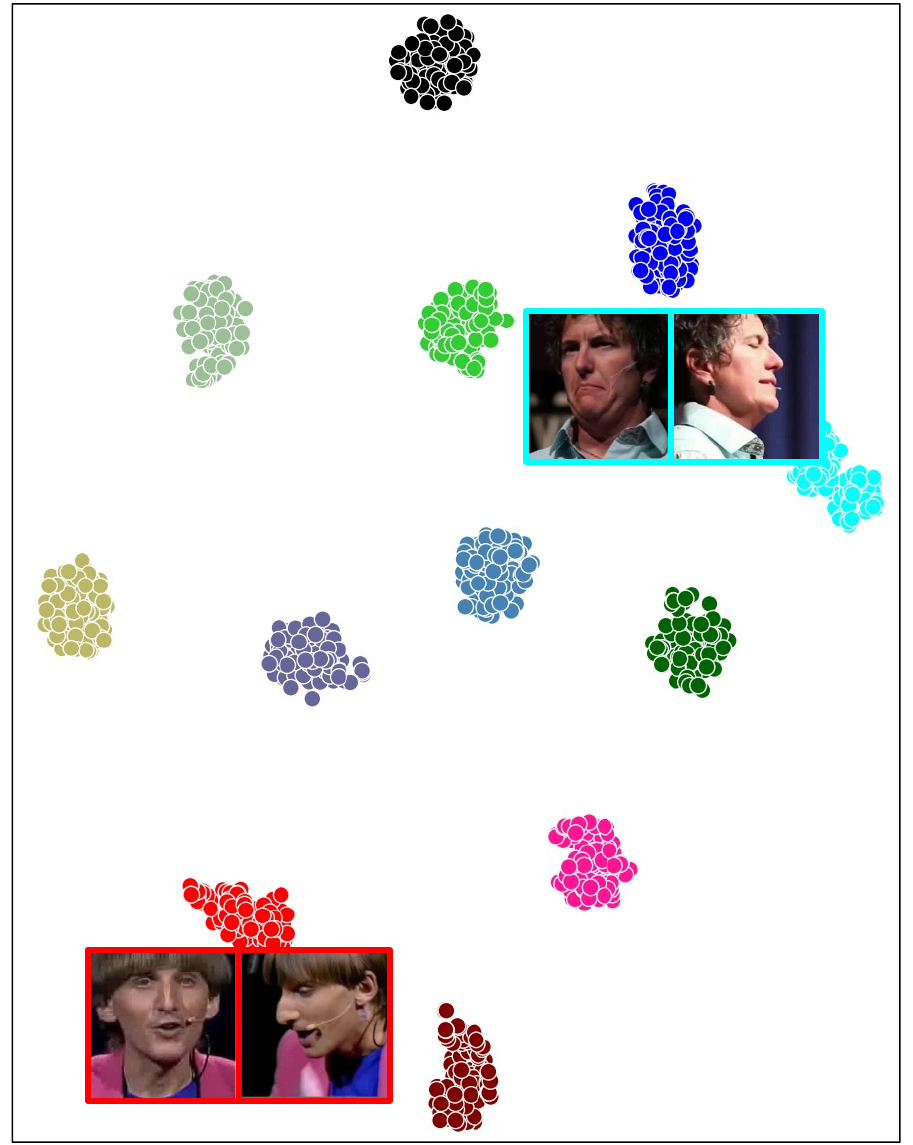}
        \label{fig:ours}
    }
    \caption{Speaker representation space of (a) Face-TTS and (b) Ours. Each colour represents a different speaker.}
    \label{fig:tsne}
\end{figure}

\newpara{The Effectiveness of Identity Features.} 
To verify the effectiveness of identity feature-based conditioning, we visualise the feature space of synthesised audio. \cref{fig:tsne} shows t-SNE~\cite{van2008visualizing} plots of x-vectors from Face-TTS and our method.
As shown in~\cref{fig:face-tts}, Face-TTS fails to cluster features derived from the same speaker. 
This implies the potential failure to generate the target voice with different styles. 
In contrast, as depicted in~\cref{fig:ours}, the proposed TTS system effectively clusters features derived from the same speaker despite the variety in head motions.
This demonstrates that our method is capable of synthesising consistent voices, even in the presence of varying motions.

\subsection{Ablation Studies}
\newpara{Analysis on Feature Aggregation in Audio Mapper.}
We perform an ablation study on the feature aggregation in our audio mapper.
w/o (energy \& $\Tilde{f}_t$) indicates the TFG model conditioned with text embedding $\boldsymbol{e}_{t}$ from the audio mapper. In this case, the TFG model can incorporate only linguistic information and it leads to our model failing to generate precise lip motions. When we additionally input the upsampled text feature $\Tilde{f}_t$ to our TFG model, the synchronisation score improves significantly. 
This is because our TTS model is optimised by reducing the prior loss between $\Tilde{f}_t$ and the target mel-spectrogram. This indicates that the $\Tilde{f}_t$ feature contains acoustic information. 
Finally, when we add the energy feature to the previous condition, our model exhibits the best performance across all metrics. This indicates that the energy of speech significantly impacts generation of detailed lip movements.

\begin{table}[!t]
    \centering
    \resizebox{0.95\linewidth}{!}
    {
        \renewcommand{\arraystretch}{1.3}
        \begin{tabular}{l|c|c|c}
        \hline\hline
        \multirow{2}{*}{\textbf{Models}} & \bf Video Quality & \bf Synchronisation & \bf Diversity \\ 
        \cline{2-4}
         &  {ID-SIM}$\uparrow$\quad & {LSE-C}$\uparrow$\quad & {DIV}$\uparrow$\quad \\ 
        \hline
        \bf{TTSF} (Ours) & \bf 0.876 & \bf 5.721 & \bf 0.143  \\ \hline
        w/o energy & 0.874 & 5.555 & 0.143 \\
        w/o (energy \& $\Tilde{f}_t$)  & 0.872 & 3.935 &  0.139 \\
        \hline\hline
        \end{tabular}
    }
\caption{Ablation study on feature aggregation in audio mapper.}
\end{table}

\section{Conclusion}
Our work introduces a unified text-driven multimodal synthesis system that exhibits robust generalisation to unseen identities. The proposed OT-CFM-based motion sampler, coupled with an auto-encoder-based noise reducer, produces realistic facial poses. Notably, our method excels in preserving essential speaker characteristics such as prosody, timbre, and accent by effectively removing motion factors from the source image. Our experiments demonstrate the superiority of our proposed method over cascade-based talking face generation approaches, underscoring the effectiveness of our unified framework in multimodal speech synthesis.

\clearpage
{
    \small
    \bibliographystyle{ieeenat_fullname}
    \bibliography{main}
}


\end{document}